\documentclass[a4paper,11pt]{article}
\usepackage[top=1in, bottom=1.25in, left=0.7in, right=0.7in]{geometry}
\usepackage{wrapfig}
\usepackage{amsmath}%
\usepackage{amsfonts}%
\usepackage{amssymb}%
\usepackage{graphicx}
\usepackage{caption}
\usepackage[font=small,labelfont=bf]{caption}
\usepackage{subcaption}
\usepackage{multirow}
\usepackage{cite}
\usepackage{url}
\usepackage{soul}
\usepackage{algorithm,algorithmic}
\usepackage{xcolor}
\usepackage{etoolbox}
\numberwithin{equation}{section}
\usepackage[hidelinks,bookmarks=false]{hyperref}
\RequirePackage{doi}
\usepackage{float}
\hypersetup{ allcolors=black}
\usepackage[bottom]{footmisc}

\newtheorem{theorem}{Theorem}

\newtheorem{remark}[theorem]{Remark}

\begin{document}
\begin{center}
{ \huge \bfseries A multi-contrast MRI approach to thalamus segmentation\\[0.4cm] }
\noindent
{Veronica Corona$^{a*}$, Jan Lellmann$^b$, Peter Nestor$^{c,d}$, Carola-Bibiane Sch{\"o}nlieb$^{a,1}$, Julio Acosta-Cabronero$^{e,f,1}$\\
\vspace{0.5cm}
\small{$^a$ Department of Applied Mathematics and Theoretical Physics, University of Cambridge, United Kingdom\\
$^b$ Institute of Mathematics and Image Computing, University of L{\"u}beck}, Germany\\
$^c$ Queensland Brain Institute, University of Queensland, Brisbane, Australia\\
$^d$ Mater Hospital, South Brisbane, Queensland, Australia\\
$^e$ Wellcome Centre for Human Neuroimaging, UCL Institute of Neurology, University College London, United Kingdom\\
$^f$ German Center for Neurodegenerative Diseases (DZNE), Magdeburg, Germany}
\end{center}
\footnotesize{$^*$ Corresponding author. Department of Applied Mathematics and Theoretical Physics, Centre for Mathematical Sciences, Wilberforce Rd, Cambridge, CB3 0WA, United Kingdom. \textit{E-mail address:} vc324@cam.ac.uk} \\
\footnotesize{$^1$ Shared senior role} \\
\newline
\small{\textbf{Keywords:} thalamus segmentation, thalamic nuclei, magnetic resonance imaging, multi-contrast segmentation}

\abstract
\noindent Thalamic alterations are relevant to many neurological disorders including Alzheimer's disease, Parkinson's disease and multiple sclerosis. Routine interventions to improve symptom severity in movement disorders, for example, often consist of surgery or deep brain stimulation to diencephalic nuclei. Therefore, accurate delineation of grey matter thalamic subregions is of the upmost clinical importance. MRI is highly appropriate for structural segmentation as it provides different views of the anatomy from a single scanning session. Though with several contrasts potentially available, it is also of increasing importance to develop new image segmentation techniques that can operate multi-spectrally. We hereby propose a new segmentation method for use with multi-modality data, which we evaluated for automated segmentation of major thalamic subnuclear groups using $T_{1}$-, $T_{2}^{*}$-weighted and quantitative susceptibility mapping (QSM) information. The proposed method consists of four steps: highly iterative image co-registration, manual segmentation on the average training-data template, supervised learning for pattern recognition, and a final convex optimisation step imposing further spatial constraints to refine the solution. This led to solutions in greater agreement with manual segmentation than the standard Morel atlas based approach. Furthermore, we show that the multi-contrast approach boosts segmentation performances. We then investigated whether prior knowledge using the training-template contours could further improve convex segmentation accuracy and robustness, which led to highly precise multi-contrast segmentations in single subjects. This approach can be extended to most 3D imaging data types and any region of interest discernible in single scans or multi-subject templates. \\

\section{Introduction}

The thalamus is composed of a complex set of sub-nuclei. It is considered the central relay station for sensory and motor information as nearly all sensory and motor signals are sent to the thalamus prior to reaching the cortex. It is also thought to have an integrative role as thalamic structures receive, process, sort and send information between specific subcortical and cortical areas, and might be involved in regulation of sleep and wakefulness, memory, emotion, consciousness, awareness, attention, ocular motility, learning and motor control processes \cite{Mai2011,Steriade.Llinas,Sherman2002}. \\
\newline
The thalamus is composed of several major substructures. The internal medullary lamina is a thin sheet of white matter that runs longitudinally through the thalamus separating it into medial and lateral regions. In the anterior part, the internal medullary lamina forks to isolate the anterior thalamic nucleus; thus, thalamic nuclei can be broken down into four regions based on their position relative to the lamina, i.e., anterior, medial, lateral and posterior subnuclear groups \cite{Chien2016,Conn2016}. 
\newline
\newline
Lesions to thalamic nuclei and their connections to the cortex can result in a wide range of neurological deficits. Thalamic alterations have been identified in several neurodegenerative diseases including Alzheimer's disease, Parkinson's disease, Huntington's disease and multiple sclerosis, the majority of which present evidence of atrophy in one or more substructures of the thalamus \cite{Steriade.Llinas,Power2015,Kassubek.etal,Amano2004}. Neurological patients also often undergo brain surgery and deep brain stimulation targeting thalamic subnuclei, thus accurate and reliable localisation of such structures are key both for research and for delivering effective clinical treatments \cite{Ondo2001,Steriade.Llinas}.
\newline
\newline
New developments in imaging techniques, including 3-7 Tesla MRI, provide greater contrast and higher spatial specificity to study the thalamus. Therefore, new strategies need to be investigated for clinical and research applications, which could potentially lead to suitable tools for predicting cognitive impairment and for monitoring disease progression in neurological patients \cite{Gringel2009}.
\newline
\newline
To date, several methods have been proposed to delineate subthalamic regions with MRI, a few of which used diffusion MRI. For example, Behrens et al. \cite{Behrens.etala} described a diffusion tensor imaging (DTI) based segmentation procedure based on coarse tractography patterns from the thalamus to the cortex, and Wiegell et al. \cite{Wiegell.etal} developed a k-means clustering algorithm to detect groups of coherent DTI-based fiber orientation. The use of the mean-shift algorithm \cite{Duan.etal} has also been proposed, whereby regional clusters and shapes are inferred from the local modes of a density estimator computed with a multivariate kernel \cite{Duan.etal}. Furthermore, Jonasson et al. \cite{Jonasson.etal} proposed a level-set method whereby a region-based force (defined using a diffusion similarity index between the most representive tensor of each level set and its neighbors) drives a set of coupled level-set functions each representing a segmented region. High angular resolution diffusion images (HARDI) have also been investigated for segmenting the thalamus. Grassi et al. \cite{Grassi.etal} proposed a k-means clustering approach whereby a specific number of initialised centroids are updated based on a weighted sum of the Euclidean distance of voxels to centroids and Frobenius distance of their orientation distribution function. Notably, however, all diffusion MRI based methods are hampered by low spatial resolution. In an attempt to overcome this limitation, Deoni et al. \cite{Deoni.etal} explored with some success the use of high-resolution quantitative MRI, namely $T_1$ and $T_2$ mapping, with a modified k-means clustering approach that combined $T_1$/$T_2$ information and center-of-mass distances to Morel atlas segmentations \cite{Morel.etal}. Further using anatomical images, Magon et al. \cite{Magon.etal} developed a method to segment thalamic subnuclei employing a voxel-wise majority vote after co-registration to multiple atlases.
\newline
\newline
Past efforts also focused on the MRI acquisition. Bender et al. \cite{Bender.etal}, for example, proposed an inversion time optimisation strategy to enhance the $T_1$-weighted contrast between gray and white matter using the 3D magnetization-prepared rapid acquisition of gradient echo (MPRAGE) sequence. Tourdias et al. \cite{Tourdias.etal} subsequently optimised MPRAGE for 7T MRI and proposed imaging at the white matter null regime both for enhancing the contrast between the thalamus and surrounding tissues and for depicting several subnuclear groups. 
\newline
\newline
Thalamus segmentation with quantitative susceptibility mapping (QSM) - a relatively new quantitative MRI contrast - has also gained increasing interest in recent times. Deistung et al. \cite{Deistung.etal} illustrated that high-resolution QSM is a superior contrast to depict thalamic substructures than $T_{2}^{*}$-weighting, the local field or $R_{2}^{*}$ maps. Therefore, considering QSM's ability to provide quantifiable information about iron content \cite{HAMETNER2018117}, that iron accumulation has been associated with several neurological disorders \cite{WARD20141045} and that thalamic lesions are not uncommon in such disorders \cite{Steriade.Llinas,Kassubek.etal}, it is highly plausible that enabling reliable segmentation of thalamic substructures could have a major impact on the study of ageing and disease.
\newline
\newline
Traditionally, however, the anatomy of the thalamus has been inferred from post-mortem tissue examinations. The most widely used histological atlas was developed by Morel et al. \cite{Krauth.etal} using an iterative approach for reconstructing the mean model from six series of maps derived from different stacks of histologically processed brains. The model, thus, is an average template incorporating topological and geometric features from only a few individuals. Morel's and other similar proprietary atlases are widely used for guiding MRI-based segmentations in neurosurgical planning, although notably, the direct superposition onto brain scans is often not fit for precision measurements, a situation often aggravated by age-related differences \cite{Steriade.Llinas}. Therefore, the development of image-guided segmentation approaches are highly relevant in this context. 

\paragraph{Our contribution.} This work proposes a new multi-contrast segmentation algorithm, and its optimisation, to exploit the full potential of $T_1$-, $T_2^*$-weighted and QSM contrast differences in thalamic subregions. 
We show that using multi-contrast information maximizes segmentation performance, by exploiting structures that become visible and enhanced in specific MR imaging protocols. In the proposed method, regions of interest defined in template space are learnt and then approximated in single subjects with spatial constraints that ensure robustness. Our multi-contrast segmentation framework can be extended to any data types and regions of interest.

\section{Methods}

The proposed semi-automated method consists of four steps: spatial normalisation, manual (reference) segmentation, pattern recognition and a final refinement step using a convex formulation.
\newline
\newline
Details on study subjects, MRI acquisition and pre-processing are given below. For now, we will assume all subject data has been spatially co-registered to a common reference space, from which multi-subject templates (one for each contrast) have been computed. We will also assume hereafter (specific details given below) that region(s) of interest has(have) been manually traced (at least once) with the aid of such templates. We then consider the following multi-class labelling problem.

\subsection{Classification}
\label{classification}

For each voxel in the image volume domain $\Omega \subset \mathbb{R}^3$, $\Omega=\{1, \dots, n_1\} \times \{1, \dots, n_2\} \times \{1, \dots, n_3\}$, we assign one of $\ell$ class labels, with each class referring to a segmented region. Let $X=\{x_i, i=1,\dots, n\}$, where $n=n_1~n_2 ~n_3$, be the vectorised volume in template space. For each $x_i$, we have $c$ image intensities or MRI parameter values, $f_1(x_i), \dots, f_c(x_i)$, one from each imaging contrast available; in this study, $T_1$-, $T_2^*$-weighted signals and QSM. We then identify a set of possible class labels, $\{0, 1, \dots, \ell-1\}$; in this particular context, we set $0$ to be the background region, $1$ the lateral thalamic subnuclear group, $2$ the medial group and $3$ the posterior group. The manual segmentation in template space is required to define the label set for the volume $X$ as $Y=\{y_i, i=1,\dots, n\}$, where $y_i \in \{0,1,2,3\}$.

\paragraph{Feature space.} In the reference coordinate system we then build the feature space: $\Psi=\{\psi_{ij}, i=1, \dots,n;$  $j=1,\dots, m\}$, assigning $x_i, \,i=1,\dots,n$, i.e., an $m$-dimensional feature vector, to each voxel. Features describe objects, in our case voxel information reflecting thalamic tissue properties. In this work we set out to develop a multi-spectral approach to exploit features from several contrast types, whereby the key features are intensity/MRI parameter values: $f_k(x_i), k=1,2,3$ from $T_1$, $T_2^*$-weighted MRI and QSM, which return different contrast characteristics for tissues with different local concentrations of water, iron, myelin, etc. For each contrast, we also selected additional features which are the result of an empirical study of the feature space. These are:  mean, $\mu$, and standard deviation, $\sigma$, across the 26-neighbourhood, and intensity/MRI parameter values for the six closest 3D neighbours in each contrast, leading to a feature space of $m=27$ dimensions. All features were then scaled by their normalised variance (i.e., with mean shifted to the origin and total variance for all features scaled to 1). It should be noted that this feature space was optimised through an investigation of classification accuracy versus feature space dimensionality on a data subset. This might differ for other data types and/or target regions.

\paragraph{Classifiers.} Let us consider the feature space $\Psi$ and the label set $Y$ for $n$ voxels. Each template voxel is therefore described by the pair $(\boldsymbol{\psi}_i, y_i)$, where $\boldsymbol{\psi}_i$ is the $m$-dimensional feature vector of voxel $x_i$ and $y_i$ is its label. We define the labeled training dataset as $ \mathcal{T}=\{(\boldsymbol{\psi}_1,y_1),...,(\boldsymbol{\psi}_n,y_n)\}$. We set out to solve a classification problem based on supervised learning, in which we train a classifier to derive a decision mapping for new observations. Initially, we explored the performance of several classification methods including Support Vector Machine, Random Forest, Naive-Bayes, k-Nearest Neighbours (\textit{k}-NN) and Parzen classifiers using a data subset. In this preliminary study, we obtained greater accuracy with two classifiers: \textit{k}-NN ($k=3$) and \textit{Parzen} classifiers, which are described in \cite{Theodoridis.Koutroumbas}. As pointed out in the context of feature selection, the optimal choice of training classifier may also vary according to data type and/or target region. 



\subsection{Convex segmentation}

Classification routines yield a posterior probability distribution $\hat{p}(u|f) \in \mathbb{R}^{n \times \ell}$ for each class, that is the probability for voxel $x$ to be assigned class $u(x)=l$ given the measured data $f(x)$. From this, winner-takes-all segmentation can be derived selecting the class with the highest probability value in each voxel. This, however, often results in scattered clusters of misclassified voxels that break the smoothness and continuity of segmented regions. Hereby, therefore, we introduce an additional convex optimisation step to further improve the spatial cohesiveness of tissue segments.
\newline
\newline
More precisely, we consider a labelling function $u: \Omega \to \mathbb{R}^\ell$ that represents the unique assignment of a label to each voxel $x$ in the image domain $\Omega$. Because this is a hard combinatorial problem, we use a convex relaxation to facilitate the optimisation, see \cite{Lellmann2011c} for an overview. 
The notion of labelling function is relaxed to $u$ taking values in the convex set defined by the unit simplex  $\Delta_\ell := \{ u \in  \mathbb{R}^n \to \mathbb{R}^\ell \,\vert\, u \geq 0, \sum_{i=1}^\ell u_i =1 \}$. Then, by choosing $J$ convex, we solve the following convex segmentation (CS) problem:

\begin{equation}
\min_{u:\Omega \to \Delta^\ell} \underbrace{\sum_{x \in \Omega} - \log \, \hat{p}\left(u(x) | f(x)\right) }_{\text{data term}}+ \underbrace{\lambda \text{TV}(u)}_{\text{regulariser}},  
\label{eq:prob_var}
\end{equation}
\newline
whereby the data term is the negative logarithm of the posterior probability distribution computed by the classifier, and the regulariser is the total variation (TV) of the labelling function $u$ defined as the L$^1$-norm of a discrete finite-difference approximation of the two-dimensional gradient of $u$. The TV regulariser on the relaxed $u$ is the convex equivalent to the length penalty on the original hardcoded labelling function and, as such, it can be thought of as introducing a penalty for long or irregular interfaces between different classes. The parameter $\lambda > 0$ balances the data term and the regulariser in the minimisation. We solve \eqref{eq:prob_var} using the fast primal-dual algorithm described in \cite{PockCremersBischofEtAl2009,ChambolleCremersPock2012}.


\subsection{Convex segmentation with additional priors}

Individual datasets are overall inferior to group-wise templates in terms of signal- and contrast-to-noise ratio. With a view then to ensure segmentation robustness at the single-subject level, we extended the forward model by the introduction of a priori information on the manual segmentation of the training template. We enabled the weighting of posterior probabilities, $\hat{p}(u | f)$, according to template-based constraint as follows:

\begin{equation}
\min_{u:\Omega \to \Delta^\ell}  \sum_{x \in \Omega} - \text{log} \big( (1-w)  \: \hat{p}(u(x)|f(x)) + w \: n \big) +\lambda \text{TV}(u) 
\label{eq:prob_var_single}
\end{equation}
\newline
where $w \in [0,1]$ is a normalised weight determining the level of prior information constraining the data term, $n:\Omega \to \{0,1\}^\ell$ is a labelled mask of thalamic subregions, and $\lambda >0$ is the regularisation parameter. 


\subsection{Study subjects}
\label{studysubjects}

Training and test datasets consisted of $\operatorname{N}=43$ (age: 59$\pm$ 7, [50-69] years old, 19 female/24 male) and $\operatorname{N}=116$ (age: 54$\pm$ 19, [20-79] years old, 56 female/60 male) healthy subjects, respectively. The latter was an aging cohort previously investigated with QSM \cite{Acosta-Cabronero2016}. All elderly subjects (age$>$50 y.o) were free of neurological or major psychiatric illness and performed normally on cognitive screening: mini-mental state examination ($\operatorname{MMSE}>27$) \cite{Folstein.etal}. 


\subsection{MRI scanning protocol}

The imaging protocol, QSM reconstruction and spatial normalization methods (briefly summarised below) are essentially identical to those in a previous aging study \cite{Acosta-Cabronero2016}. \\
\newline
All participants were scanned on a Siemens Verio 3 Tesla MRI system with a 32-channel head coil (Siemens Medical Systems, Erlangen, Germany) under the following imaging protocol: 
\newline
$T_2$-weighed fast spin echo images were acquired and visually inspected to ensure vascular pathology was not significant in any subject as for standard clinical routine. Scan parameters were as follows: flip angle ($\alpha$)/ echo time (TE)/ receiver bandwidth (RB)/ turbo factor/ number of echo trains per slice/ echo spacing/ repetition time (TR) = 150$^{\circ}$/ 96 ms/ 220 Hz per pixel/ 18/ 13/ 9.64 ms/ 8160 ms; matrix, $320\times320$ (in-plane resolution: $0.7\times0.7 \text{ mm}^2$); 45 axial slices for whole-brain coverage (thickness: 3 mm; gap: 0.9 mm); two-fold parallel acelleration was enabled for phase encoding giving a total scan time of 1:56 minutes.\\
\newline
$T_1$-weighed data were acquired using a 3D MPRAGE sequence \cite{MuglerIII1990} with the following scan parameters: inversion time/ $\alpha$/ TE/ RB/ echo spacing/ TR = 1100 ms/ 7$^{\circ}$/ 4.37 ms/ 140 Hz per pixel/ 11.1 ms/ 2500 ms; $256\times256\times192$ matrix dimensions (straight-sagittal orientation), $1\times1\times1$ $\text{mm}^3$ voxel size, two-fold parallel acceleration and further 7/8 partial Fourier undersampling for phase encoding. The total scan time was 5:08 minutes.\\
\newline
$T_2^*$-/susceptibility-weighted data were obtained from a fully flow-compensated, 3D spoiled gradient-echo sequence. Scan parameters were: $\alpha$/ TE/ RB/ TR = 17$^{\circ}$/ 20 ms/ 100 Hz per pixel/ 28 ms; matrix, $256\times224\times80$ with voxel resolution of $1\times1\times2$ $\text{mm}^3$; and two-fold parallel acceleration for phase encoding. The total scan time was 5:32 minutes. All susceptibility maps were inspected to exclude subjects with severe calcifications or extensive haemorrhages.\\

\subsection{MRI data pre-processing}

\subsubsection{QSM reconstruction}

QSM is a relatively new contrast modulated by the local content of chemical species that have different magnetic susceptibilities than soft tissue and water \cite{Wang.Liu}. Myelin phospholipids and calcifications, for example, are more diamagnetic than water; whereas, iron - the most abundant transition metal in the human brain and the dominant source of QSM contrast - is greatly paramagnetic \cite{HAMETNER2018117}. Specific details on susceptibility reconstruction can be found elsewhere \cite{Acosta-Cabronero2016}.\\

\subsubsection{Spatial standardisation}

Radio-frequency (RF) bias corrected \cite{Tustison.etal} $T_2^*$-weighted magnitude images were affine co-registered to their corresponding bias-corrected MPRAGE volume using ANTs (\url{http://stnava.github.io/ANTs/}) \cite{ants}. Subsequently, all bias-corrected anatomical $T_1$-weigthed MPRAGE images were used to generate a study-wise space using a previously described ANTs routine \cite{Avants.etal,Acosta-Cabronero2016}. Finally, all $T_2^*$-weighted images and susceptibility maps were normalised to the same coordinate system through the warp composition of the above transformations and high-order interpolation.

\subsubsection{Manual (reference) segmentation}

\begin{figure}[h!]
	\centering
		\begin{subfigure}[b]{\textwidth}
\includegraphics[width=0.3\textwidth]{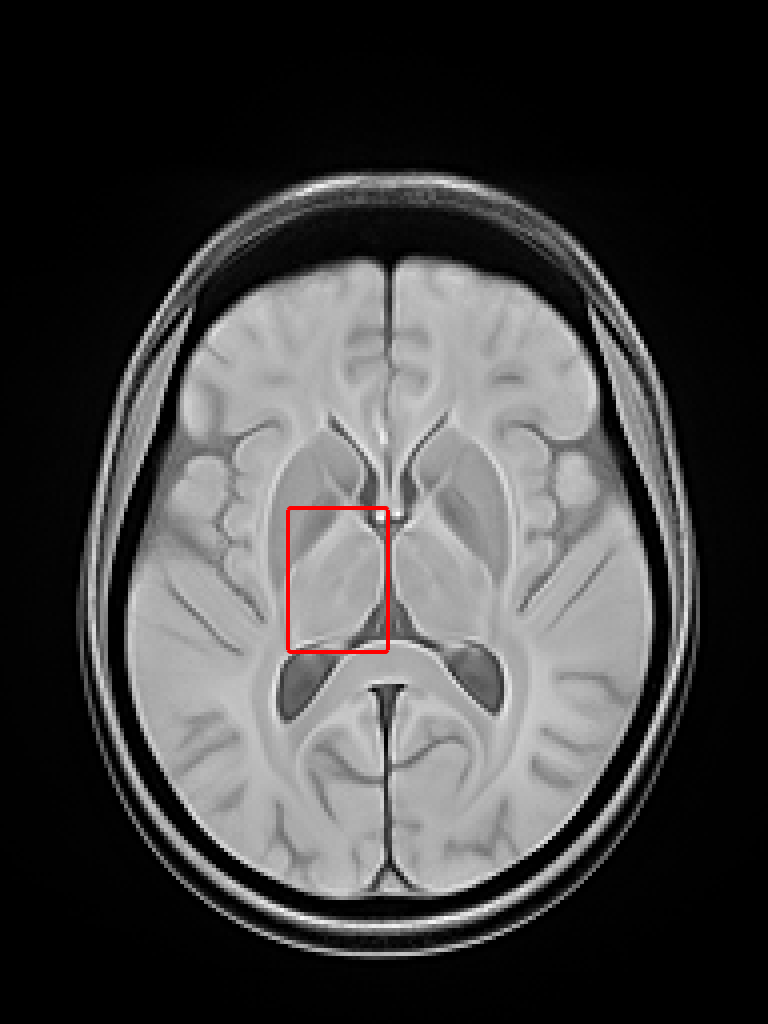}
\includegraphics[width=0.3\textwidth]{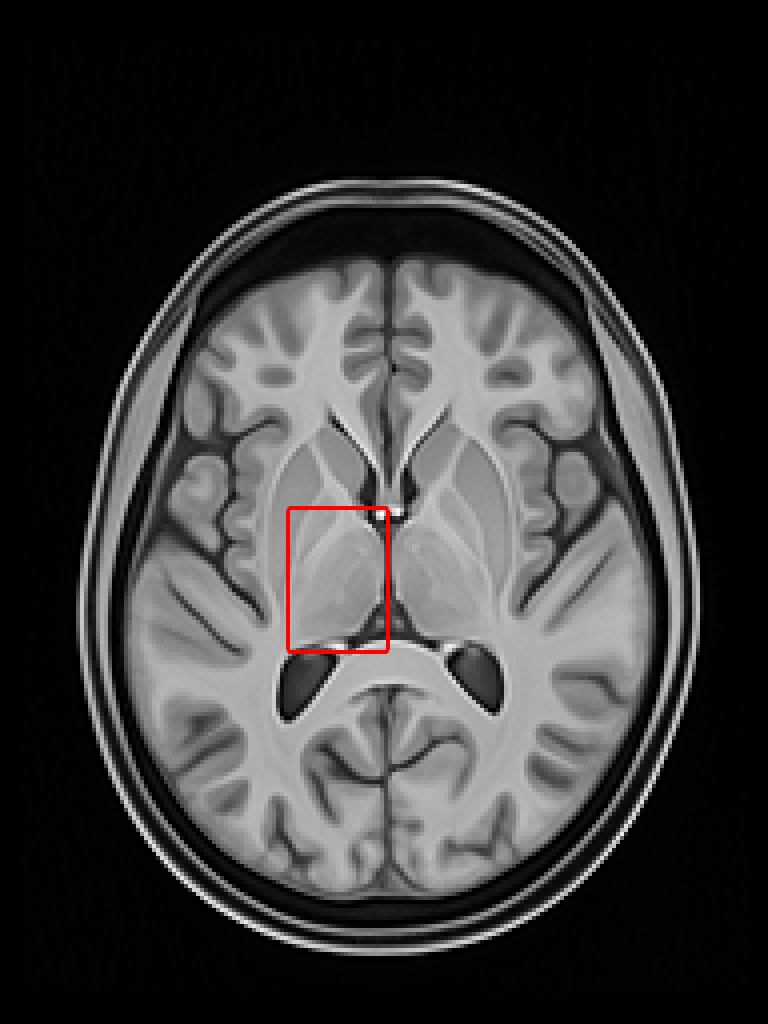}
\includegraphics[width=0.3\textwidth]{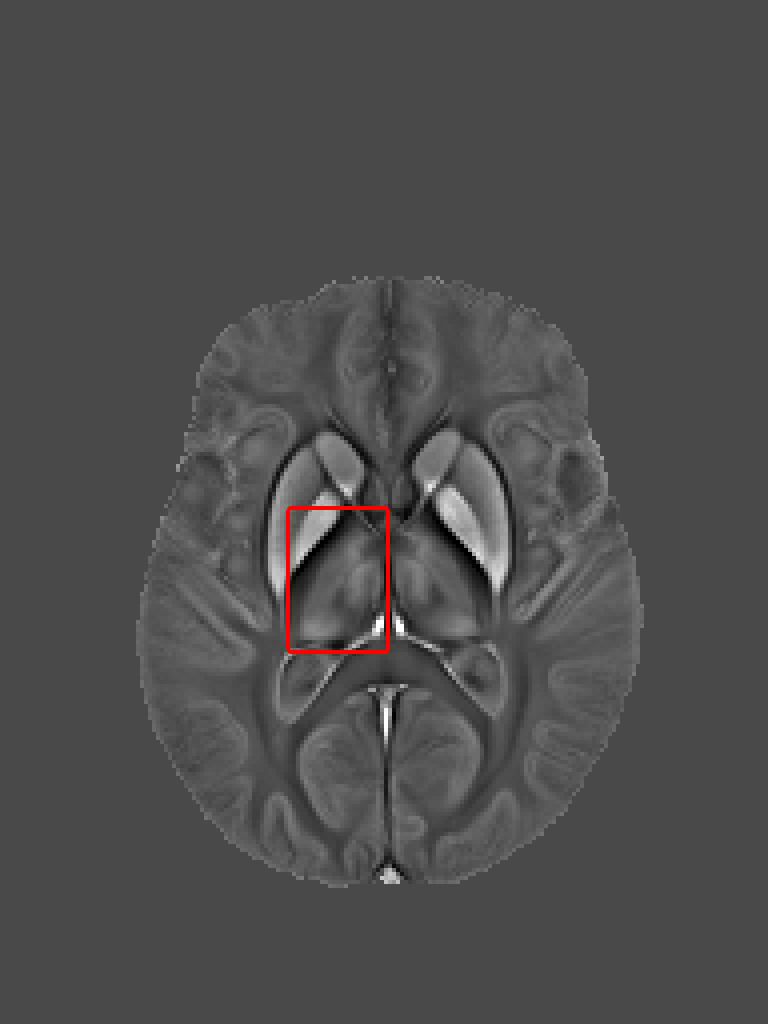}

	\caption{Representative axial cuts (from left to right) of $T_2^*$-, $T_1$-weighted and QSM templates for the $N=116$ test dataset.}
	\label{fig:contrasts}
	\end{subfigure}
	\begin{subfigure}[b]{\textwidth}
	\centering{
	\includegraphics[width=0.25\textwidth]{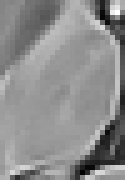}
	\includegraphics[width=0.25\textwidth]{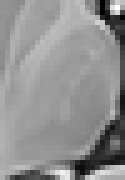}
	\includegraphics[width=0.25\textwidth]{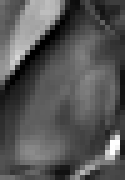}}
	\caption[Magnified views of the left thalamus for the three templates.]{Magnified views of the left thalamus for the three templates.}
	\label{fig:tmq1}
	\end{subfigure}
	\caption{Anatomical detail from group-average MRI templates.}
	\label{fig:tmq}
\end{figure}

\begin{figure}[h!]
        \centering
        \begin{subfigure}{.50\textwidth}
                    \includegraphics[width=\textwidth]{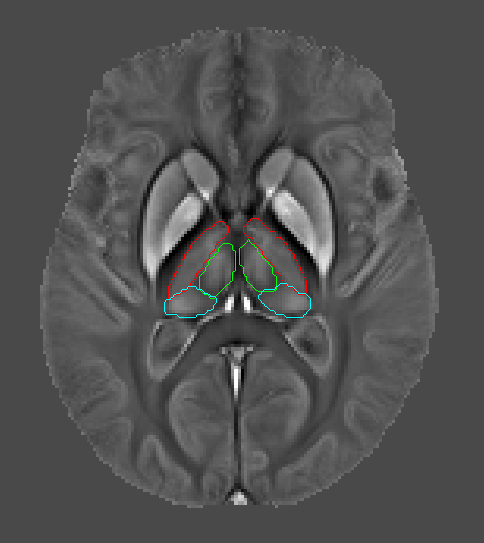}
          \caption{Manual contour overlays onto the $N$=116 average QSM template. \newline}
        \end{subfigure}%
        \hspace{10mm}%
        \begin{subfigure}{.2\textwidth}
        \vspace{-4mm}
          \includegraphics[width=0.9\textwidth]{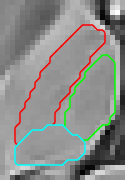}
	\caption{$\operatorname{T}_2^*$-weighting}
          \includegraphics[width=0.9\textwidth]{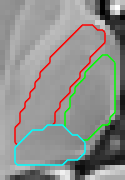}
          \caption{$T_1$-weighting}
        \end{subfigure}
        \caption{Manual segmentation of three major thalamic subnuclear groups. Left: Bilateral contours overlaid onto the $N$=116 QSM template. Red contours denote the lateral nuclear group, green contours correspond to the medial group, and cyan contours illustrate the posterior group segmentation. Right: Magnified view of the left thalamus showing manual contours overlaid onto $T_2^*$- and $T_1$-weighted templates.}
        \label{fig:manual}
      \end{figure}

Three templates were subsequently obtained from averaging $T_1$-, $T_2^*$-weighted and QSM data across subjects in the study-wise space (\autoref{fig:tmq}). This was performed separately for training and testing data.
\newline
\newline
Then, three major thalamic subregions, namely lateral, medial and posterior nuclear groups, were manually traced as illustrated in \autoref{fig:manual}. The manual annotations from the training-average template were utilised in the training phase of the algorithm as ground truth. In addition, thalamic subregions from the average test template ($N=116$) and for $N=6$ individual test datasets were delineated for algorithm validation (see next subsection).

\subsection{Performance evaluation} 

Segmentation performance was assessed through visualisation of the confusion matrix (incorporating exact error distributions). For simplicity, however, in this study we report two representative summary measures: the global classification error rate (i.e., the overall proportion of erroneously classified voxels) and the true positive (TP) rate for non-background (i.e., subnuclear group) regions. At the outset, general performance was evaluated on the high-contrast $N$=116 template dataset, which included a comparison with standard Morel atlas based segmentation. Subsequently, error measures were also computed for individual (noisier) test data.

\subsection{Methods summary}

In Algorithm \autoref{alg1}, we summarise the proposed methodology for multi-contrast MRI segmentation. The first stage of the algorithm trains a classifier for use in stage two. Given then a "new" multi-contrast MRI dataset to be segmented, all contrast images must be first realigned to a common space, then the "new subject" segmentation pipeline can be applied as follows:
\newline
\begin{itemize}
\item \textbf{supervised learning (testing)}, given the trained classifier, its mapping is applied to independent test data yielding class labels and posterior probabilities.
\item \textbf{multi-class convex segmentation}, where posterior probabilities are used in the data term of the convex optimisation formulation defined in \autoref{eq:prob_var}.
\end{itemize}

The supervised learning and convex segmentation steps of the algorithm were implemented in MATLAB R2017b (The Mathworks Inc., Natick, MA, USA) and are available at\\ \url{https://github.com/veronicacorona/multicontrastSegmentation.git}.

\begin{algorithm}[t]
\begin{algorithmic}
\STATE{ \textbf{Training stage}}
\STATE{ \textbf{Input:} Multi-contrast training data}
\STATE{\qquad\textbf{1:}} Spatial normalisation
\STATE{\qquad\textbf{2:}} Contrast-specific template generation
\STATE{\qquad\textbf{3:}} Manual (or atlas based) template segmentation
\STATE{\qquad \textbf{4:}} Supervised learning (training) 
\STATE{\textbf{Output:} Trained classifier}
\newline
\STATE{ \textbf{New segmentation}}
\STATE{ \textbf{Inputs:} Multi-contrast test data and a trained classifier}
\STATE{\qquad \textbf{5:}} Spatial normalisation
\STATE{ \qquad\textbf{6:}} Supervised learning (testing) 
\STATE{ \qquad\textbf{7:}} Multi-contrast convex segmentation
\STATE{\textbf{Output:}} Regional contours
\end{algorithmic}
\caption{Procedural steps for multi-contrast segmentation.}
\label{alg1}
\end{algorithm}

{\remark All MRI datasets in this study were spatially standardised via nonlinear co-registration to a common coordinate system. This enabled custom training from a single set of regional contours in template space. Future applications of this algorithm could alternatively consider using manual tracings from each individual training dataset. The only requirement is that all contrasts for a given subject must share a common frame of reference.}

\section{Results} 

In what follows we present our numerical results obtained independently for the $\operatorname{N}=116$ test dataset described in \autoref{studysubjects}.

\subsection{Qualitative assessment}

\begin{figure}[h!]
	\centering
\begin{subfigure}[b]{0.2\textwidth}
\includegraphics[width=\textwidth]{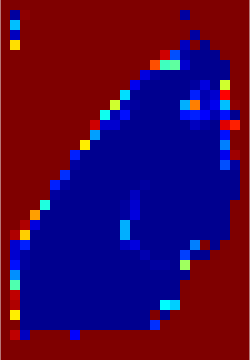}
\caption{Background}
\label{fig:p1}
\end{subfigure}
\begin{subfigure}[b]{0.2\textwidth}
\includegraphics[width=\textwidth]{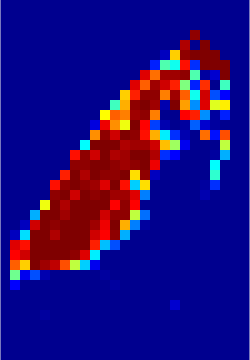}
\caption{Lateral}
\label{fig:p2}
\end{subfigure}
\begin{subfigure}[b]{0.2\textwidth}
\includegraphics[width=\textwidth]{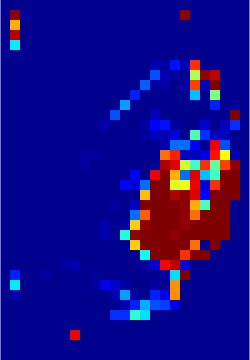}
\caption{Medial}
\label{fig:p3}
\end{subfigure}
\begin{subfigure}[b]{0.27\textwidth}
\includegraphics[width=\textwidth, height=5.05cm]{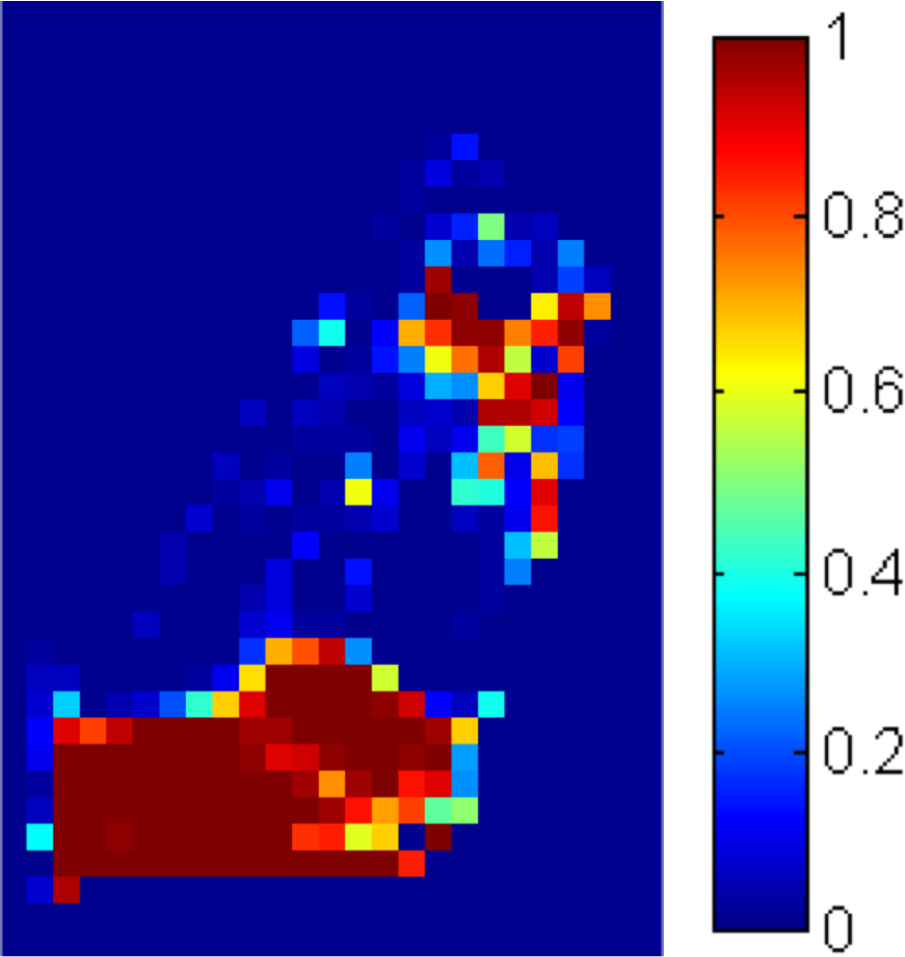}
\caption{Posterior}
\label{fig:p4}
\end{subfigure}

		\begin{subfigure}[b]{0.19\textwidth}
\centering{
\includegraphics[width=\textwidth]{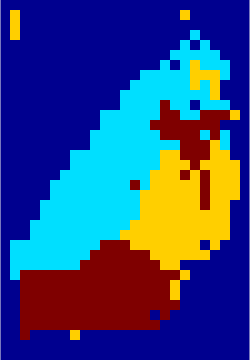}}
\caption{Classification}
\label{fig:labclass}
\end{subfigure}~
\begin{subfigure}[b]{0.19\textwidth}
\centering{
\includegraphics[width=\textwidth]{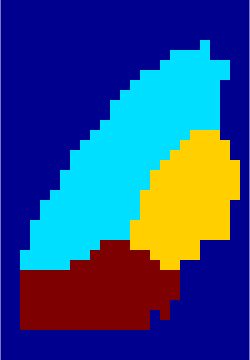}}
\caption{CS}
\label{fig:labconv}
\end{subfigure}
	\caption[Posterior probability maps]{Top row: posterior probabilities of the four thalamic tissue classes for the $N$=116 template dataset obtained with Parzen classification. Bottom row: (left) labels derived from the posterior probabilities, and (right) refined segmentation using convex segmentation (CS) on the posterior map.}
	\label{fig:posteriormaps1}
\end{figure}

In this implementation, classifiers were set out to assign four posterior probabilities per voxel, i.e., those of belonging to background, lateral, medial and posterior subregions of the thalamus. \autoref{fig:p1}-\ref{fig:p4} shows posterior probability maps using the Parzen classifier on $N=116$ average-template data. The figure indicates that accurate classification of specific subnuclear groups and the background region is feasible; supporting, thus, the choice of feature space and classifier. Overall, the best performing algorithms in our preliminary assessment were \textit{k}-NN and Parzen classifiers. For k-NN, the optimal number of nearest neighbours, $k$, was $k=3$. For the Parzen classifier, the empirically optimal parameter $h$, i.e., the width of the Gaussian smoothing kernel, was $h=0.1668$. \autoref{fig:labclass} further confirmed that the Parzen classifier output is overall in agreement with a priori knowledge on the regional distribution of subnuclear groups. However, as predicted, winner-takes-all local classification introduced undesirable regional discontinuities. This was substantially mitigated through the additional CS step as shown in \autoref{fig:labconv}.
\newline

\subsection{General performance evaluation}

\subsubsection{Convex segmentation validation}

\begin{figure}[h!]
	\centering
		\includegraphics[width=0.5\textwidth]{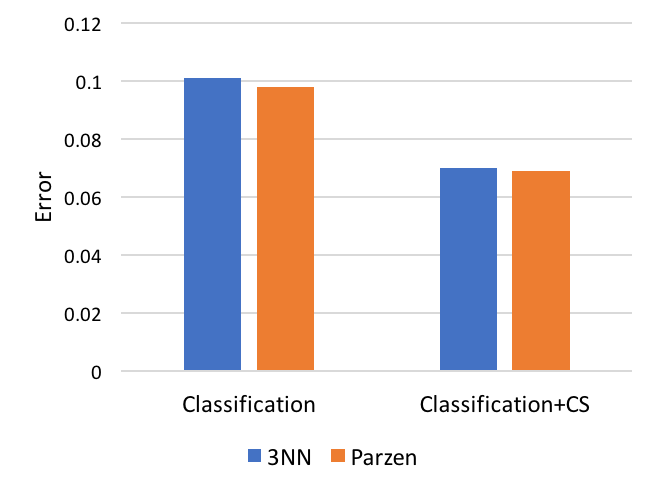}
	\caption[Error rates]{Error rates for (left) classification and (right) classification followed by convex segmentation (CS) on $N$=116 template data. Bars represent misclassification frequency, i.e., overall proportion of errors relative to the manually traced ground truth.}
	\label{fig:bar_mean_convex}
\end{figure}

Indeed, as shown in \autoref{fig:bar_mean_convex}, the introduction of convex segmentation systematically improved classification performance, which, in turn, also confirmed that posterior probability maps from both \textit{k}-NN and Parzen classifiers are suitable pre-conditioners for the CS formulation in \eqref{eq:prob_var}.
\newline

\subsubsection{Algorithm comparison}

\begin{figure}
\centering
\begin{subfigure}[b]{0.3\textwidth}
\includegraphics[width=\textwidth]{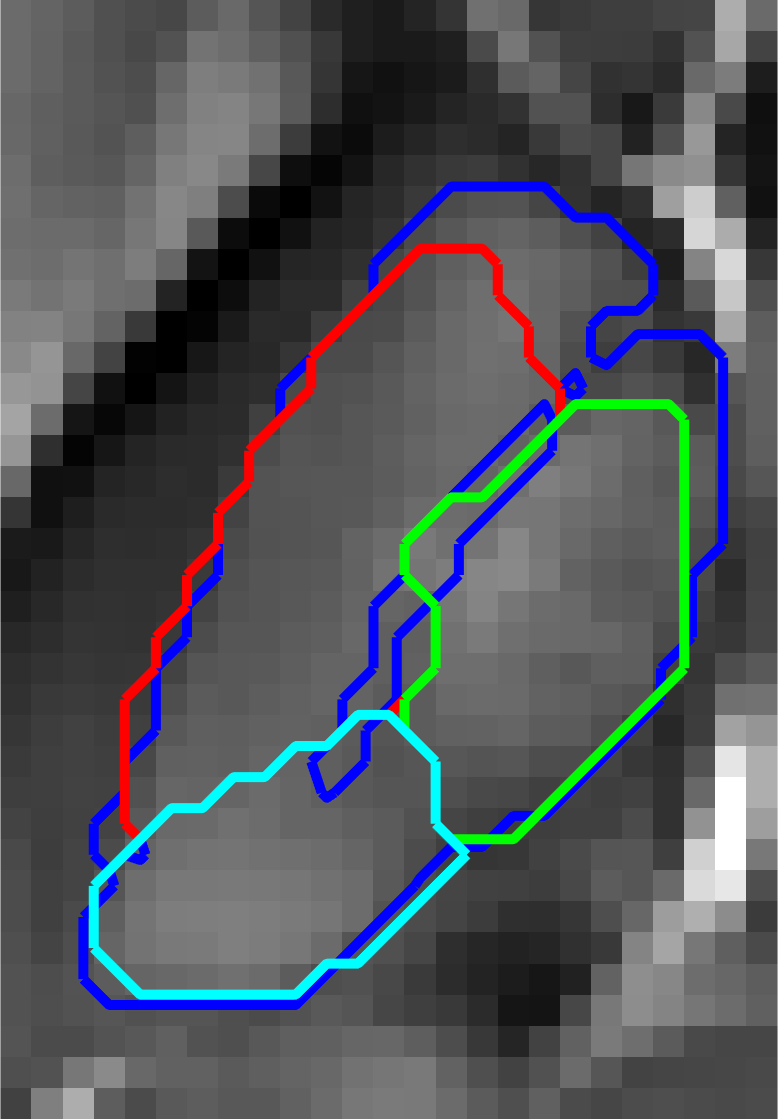}
\caption{3NN+CS}
\end{subfigure}~
\begin{subfigure}[b]{0.3\textwidth}
\includegraphics[width=\textwidth]{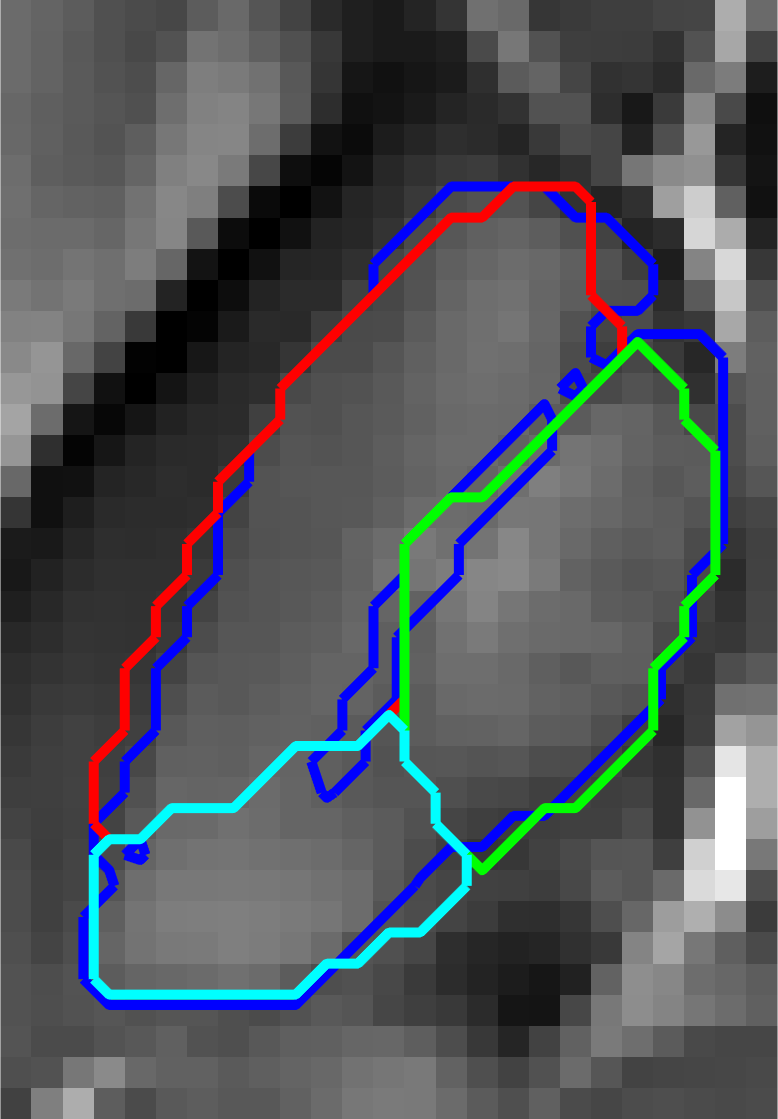}
\caption{Parzen+CS}
\end{subfigure}~
\begin{subfigure}[b]{0.3\textwidth}
\includegraphics[width=\textwidth]{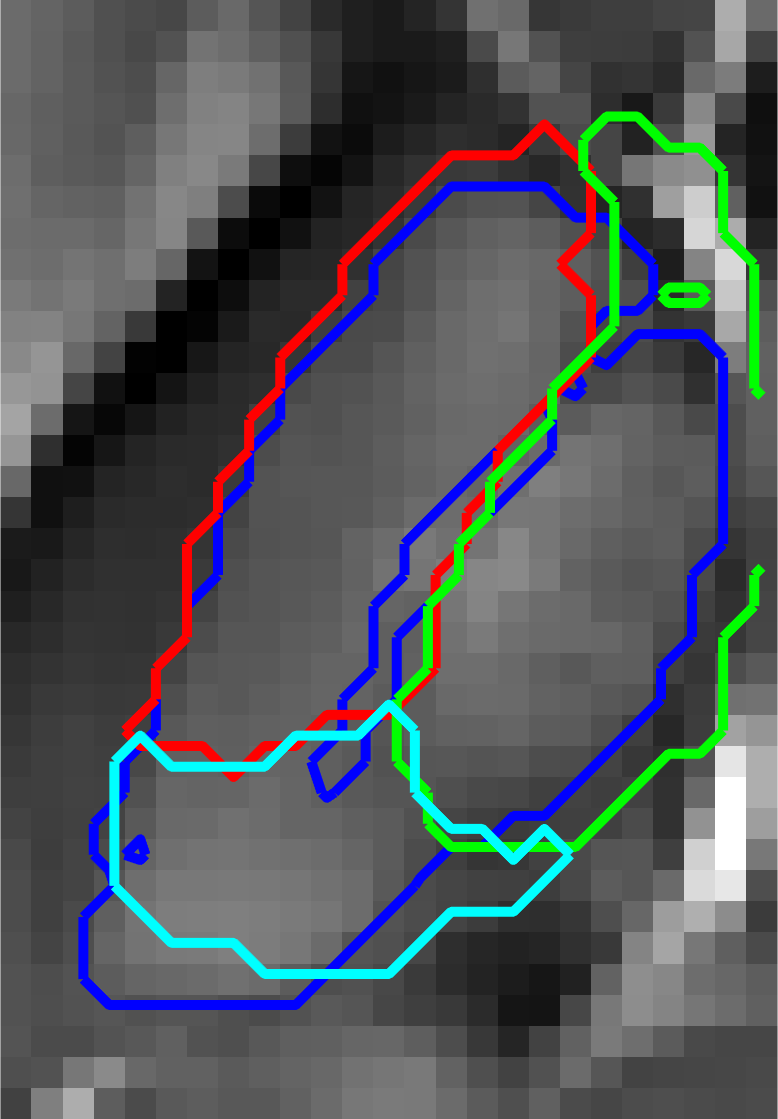}
\caption{Morel atlas}
\end{subfigure}
\caption{Convex segmentation results (for the $N$=116 template dataset) from different algorithmic implementations and the Morel atlas overlay onto the $N=116$ QSM template. The blue overlay represents the ground truth's outer contour. Red, green and cyan contours are the results for the the different approaches.}
\label{fig:res1}
\end{figure}

\autoref{fig:res1} illustrates segmentation results for all methods herein evaluated. Outputs from the proposed multi-contrast method were in greater agreement with the manual ground truth than atlas-based Morel segmentation, which is solely based on template co-registration.
\newline
\newline
It is worth noting that in this particular implementation the background region outsizes (approximately 4:1) the extent of putative thalamic regions. Therefore, segmentation results are reported in \autoref{tab1} both as global classification errors and true positive rates; the latter computed for non-background regions only. Such an evaluation confirmed the proposed algorithm outperforms Morel atlas segmentation on all performance metrics. Pre-conditioning with 3-NN and Parzen based posterior probabilities minimised classification error and true positive rates, respectively.
\newline
\begin{table}[h!]
\centering
\begin{tabular}{| c | c | c | c |}
\hline
\textbf{Classifiers} & \textbf{\% global error} & {\textbf{\% TP (nuclei)}} \\
\hline
3-NN+CS & 7.0 & 74.8\\ 
\hline
Parzen+CS & \textbf{6.9} & \textbf{88.4} \\ 
\hline
Morel Atlas & 13.3 & 69.7\\ 
\hline
\end{tabular}
\caption[Algorithm performance]{Segmentation performance for the new algorithmic implementations and the standard Morel method applied to the N=116 template dataset. The proposed implementations outperformed standard Morel segmentation on both performance metrics: global error and true positive (TP) rate.}
\label{tab1}
\end{table}

\subsubsection{Regularisation parameter selection for convex segmentation}

The 3-NN and Parzen based segmentation results in \autoref{fig:res1} and \autoref{tab1} were obtained through solving the convex optimisation problem defined in \eqref{eq:prob_var}, which has a regularisation multiplier, $\lambda$, that requires optimisation for optimal solution smoothness. In this study, $\lambda$ was optimised empirically on a small subset: for 3-NN, we chose $\lambda=1$, and for Parzen $\lambda=5$. We then confirmed the validity of this choice calculating overall classification errors (on the $N=116$ template dataset) for a range of regularisation parameters. Results from this validation experiment are summarised in \autoref{tab4}.
\newline 

\begin{table}[h!]
	\centering
		\begin{tabular}{| c | c | c | c |}
			\hline
  \multicolumn{2}{| c |}{\centering \textbf{3-NN}} &
	\multicolumn{2}{| c |}{\centering \textbf{Parzen}}\\
\hline
$\lambda$ & {\% error}&$\lambda$ & {\% error} \\
\hline
0.1 & 9.5 & 4 & 7.3\\
0.5 & 7.0&4.5 & 7.2 \\
\textbf{1} & 7.0  & \textbf{5} & 6.9 \\
1.5 &10.0&5.5 & 6.9\\
\hline
		\end{tabular}
	\caption{Classification error as a function of classifier and $\lambda$ parameter.}
	\label{tab4}
\end{table}

\subsubsection{Algorithm performance as a function of input data}

\begin{figure}[h!]
	\centering
\includegraphics[width=0.95\textwidth]{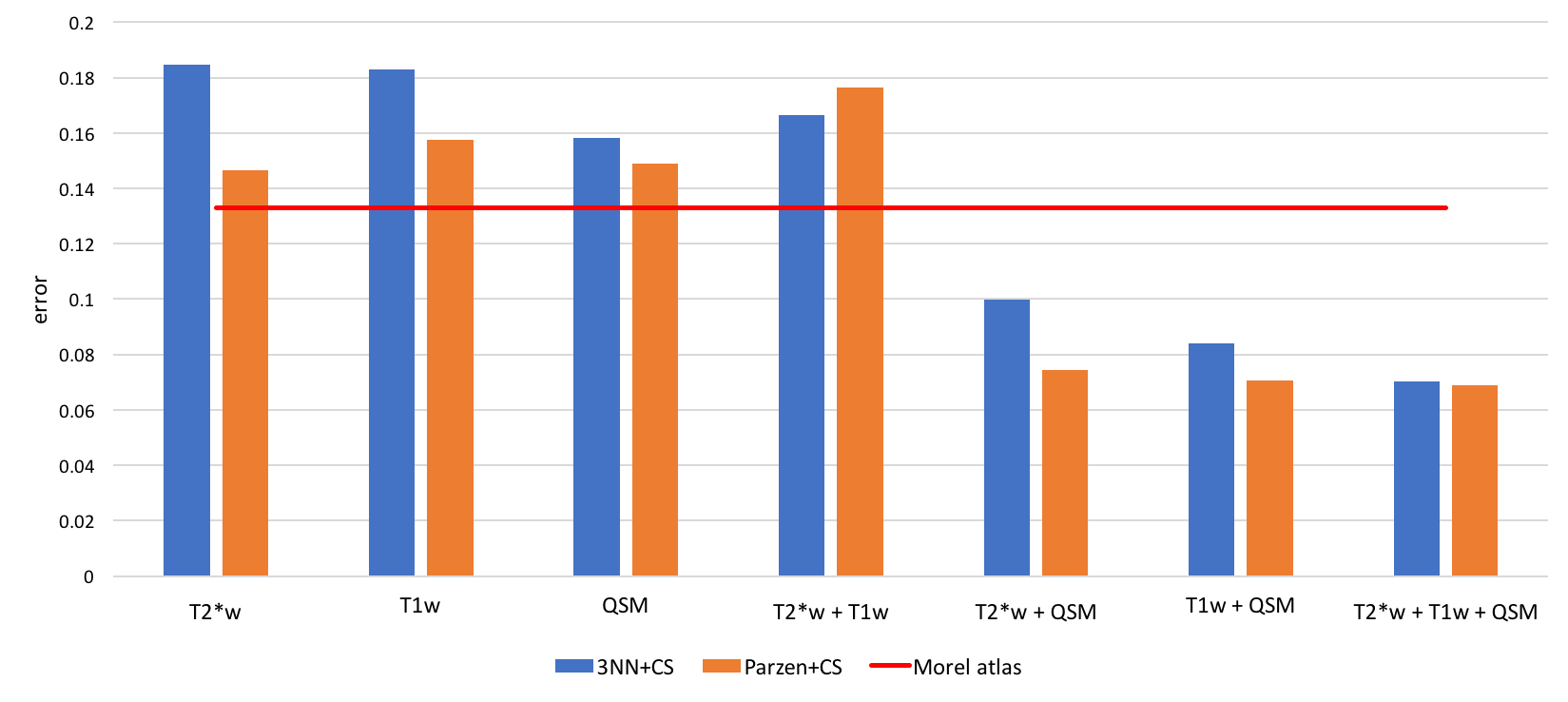}
	\caption[Algorithm performance as a function of input data]{Algorithm performance comparison on segmentation of the $N$=116 template thalamus as a function of input MRI data. Classification errors were greatly reduced when combining QSM with other contrasts. The global minimum error was obtained using all three contrasts. }
	\label{fig:onecontrast}
\end{figure}

A unique aspect of the proposed algorithm is that it can integrate information from several MRI contrasts capturing simultaneously different views of the anatomy. In this study, we hypothesised that $T_1$-, $T_2^*$-weighting and QSM all provide differentially relevant information about internal thalamic boundaries. In order to substantiate this claim, we estimated algorithm performance for all the available combinations, i.e., one, two or three data types, using the same $27$-dimensional feature space that was previously optimised. CS errors are shown in \autoref{fig:onecontrast} for pre-conditioning with both 3-NN and Parzen classifiers. Interestingly, using single contrasts alone as input data led to sytematically greater error rates than when using QSM in combination with other contrast types. Confirming our hypothesis, the best segmentation performance was obtained when using all MRI information. Although some differences were observed, overall both pre-conditioning approaches, i.e., 3-NN and Parzen classification, yielded relatively similar error rates throughout.
\newline

\subsection{Convex segmentation with additional priors for increased performance in single subjects}

\begin{figure}[h!]
\centering
\begin{subfigure}[b]{0.4\textwidth}
\includegraphics[width=\textwidth]{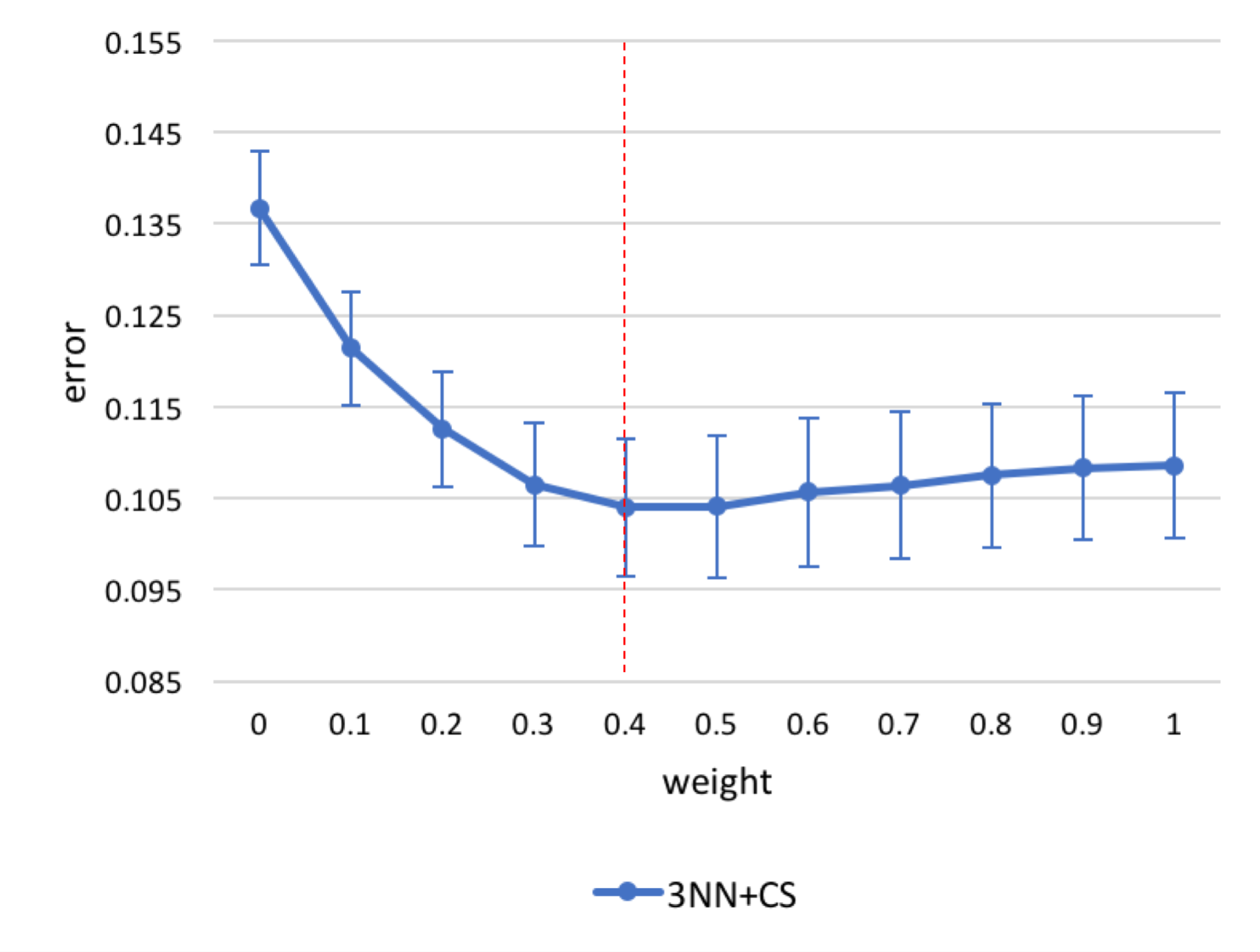}
\caption{3-NN}
\end{subfigure}
~
\begin{subfigure}[b]{0.4\textwidth}
\includegraphics[width=\textwidth]{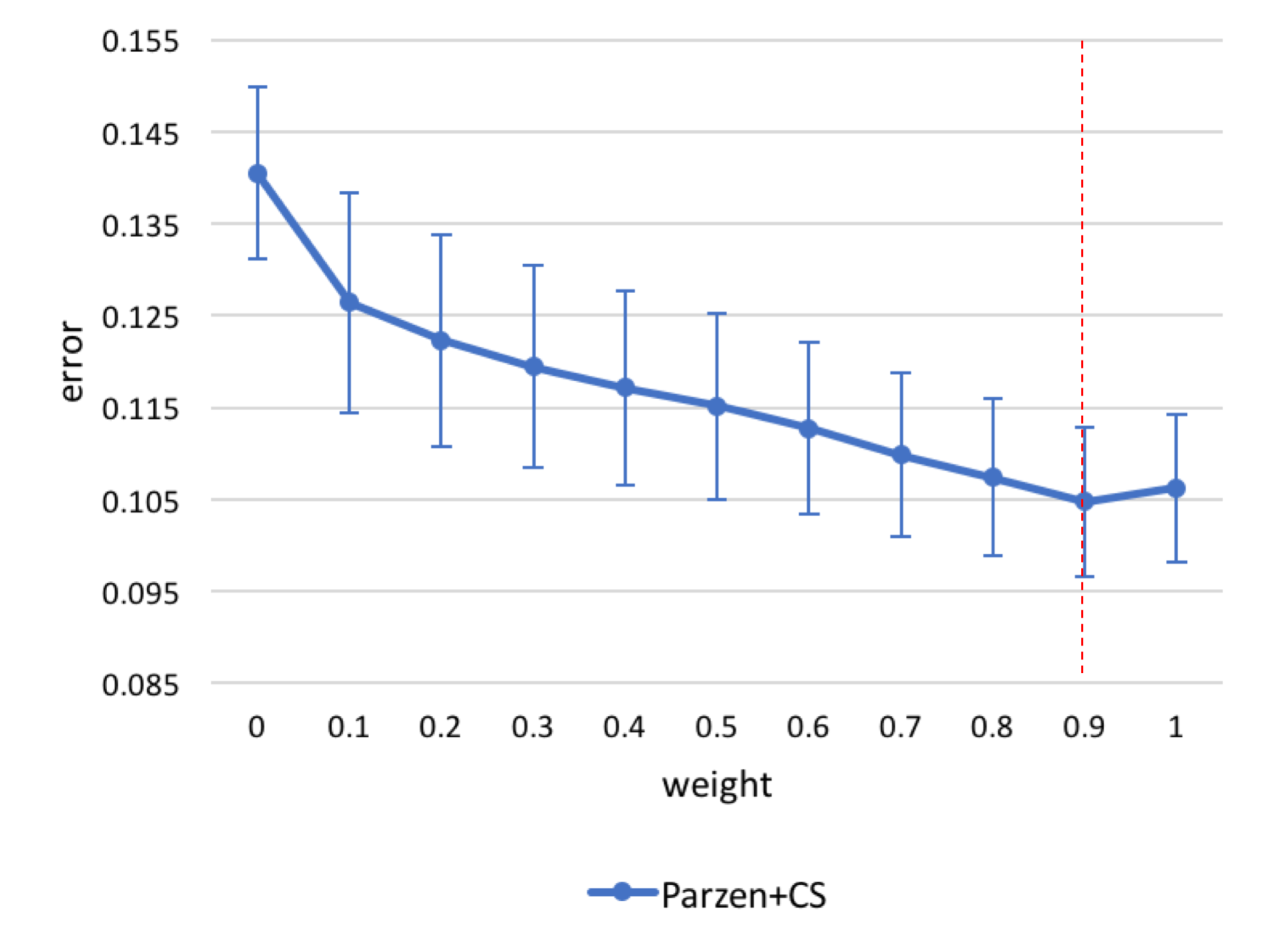}
\caption{Parzen}
\end{subfigure}
\caption{Classification error (with respect to manual gold standard) as a function of $w$. Data points and error bars denote mean and S.E.M. across $N=6$ test datasets. The red lines indicate the empirical optima $w$ ($w=0.4$ for 3-NN and $w=0.9$ for Parzen).}
\label{fig:w_err}
\end{figure}

We also confirmed that constraining the data term for fidelity with training-average tissue priors is feasible and desirable to improve accuracy and robustness in single-subject thalamic subnuclei segmentation. The consistency weight, $w$ in \eqref{eq:prob_var_single}, represents a trade-off between the calculated posterior probabilities and template-based priors. On a $N=6$ test subset, we explored how classification error varies as a function of $w$. This is illustrated in \autoref{fig:w_err}, which indicated that 3-NN is generally preferred (to Parzen based training) in this context; 3-NN pre-conditioning resulted in greater accuracy with optimal performance for a critical $w=0.4$. 
\newline

\begin{figure}[h!]
\centering
\begin{subfigure}[b]{\textwidth}
\centering
\includegraphics[width=0.25\textwidth]{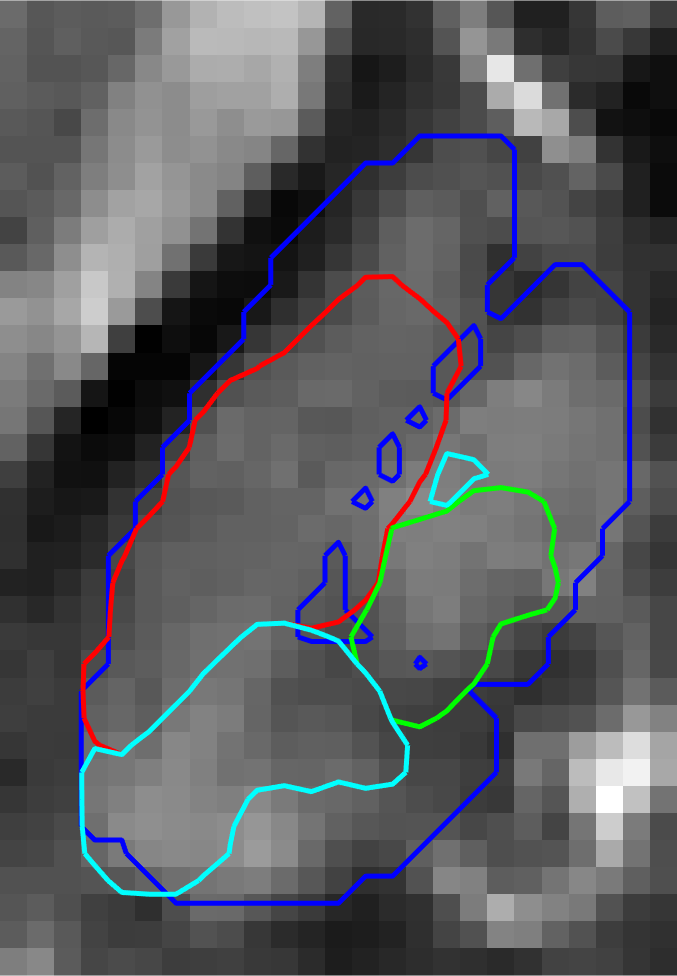}
\includegraphics[width=0.25\textwidth]{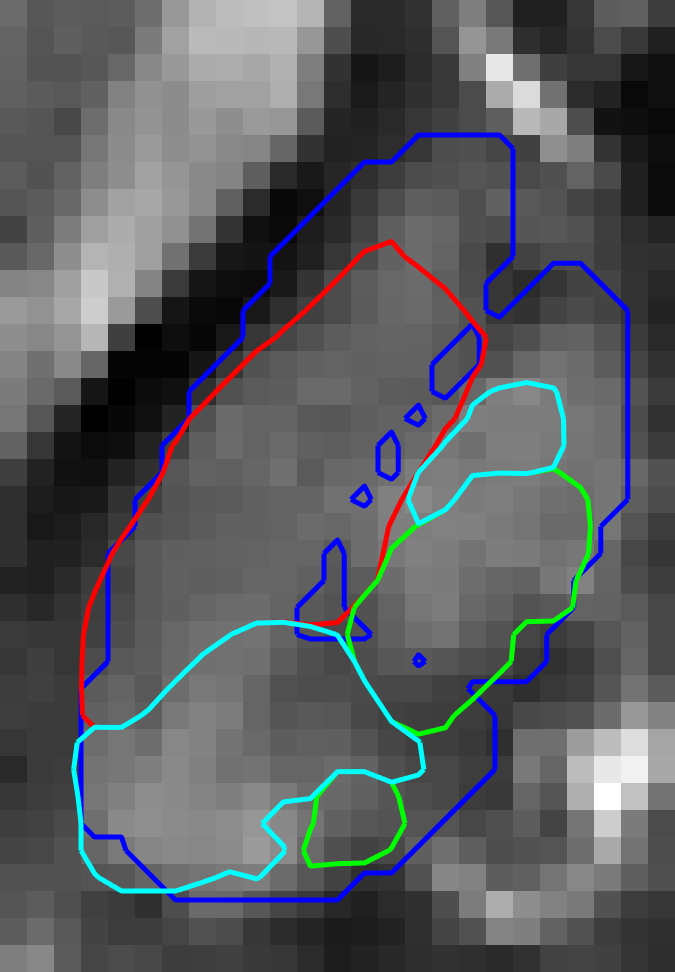}
\caption{Convex segmentation with (left) 3-NN and (right) Parzen based pre-conditioners.}
\end{subfigure}
\begin{subfigure}[b]{\textwidth}
\centering
\includegraphics[width=0.25\textwidth]{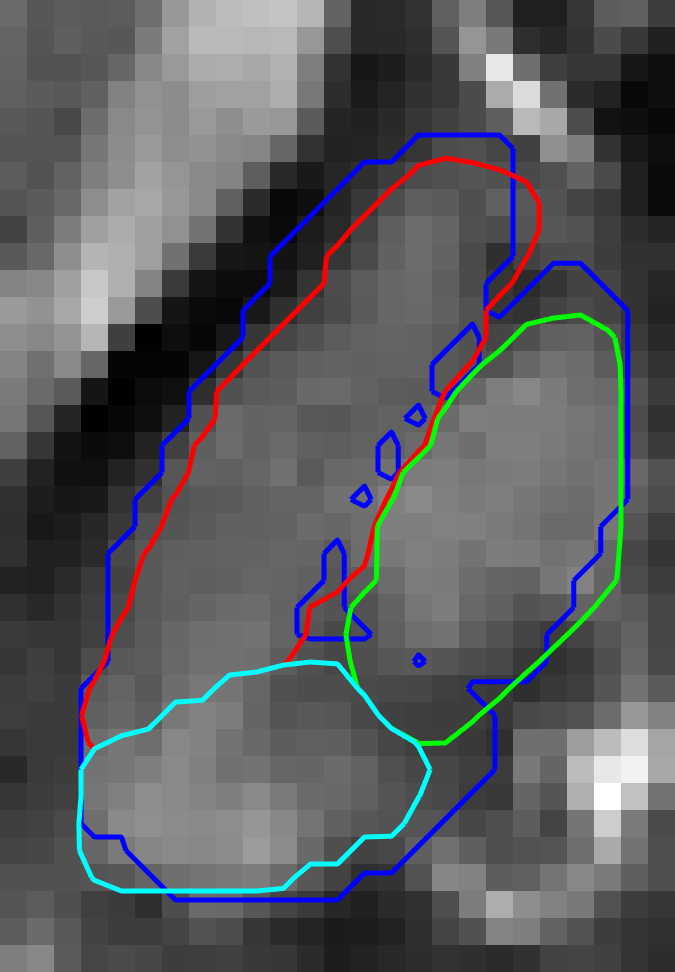}
\includegraphics[width=0.25\textwidth]{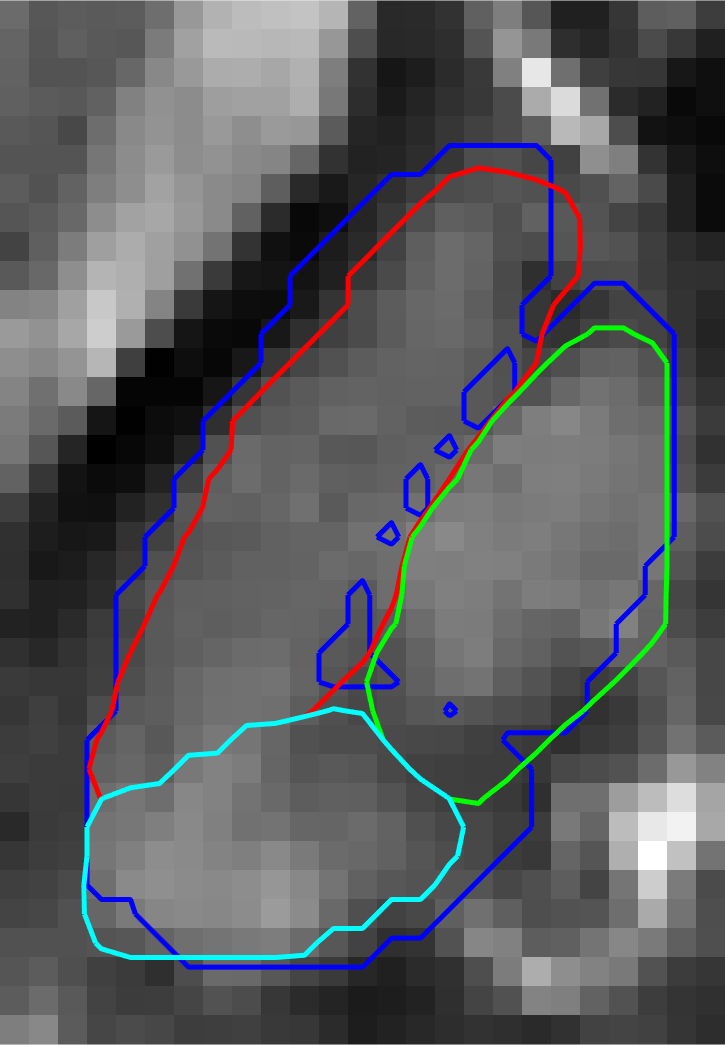}
\caption{Weighted convex segmentation with 3-NN (left, 40 \% prior) and Parzen based pre-conditioners (right, 90 \% prior).}
\label{fig:s7weight}
\end{subfigure}
\caption{Representative convex segmentation for single-subject data with and without training-average priors. The blue overlay represents the ground truth's outer boundary. Red, green and cyan contours are the results for the different approaches.}
\label{fig:single10}
\end{figure}

\begin{figure}[h!]
	\centering
		\includegraphics[width=1.00\textwidth]{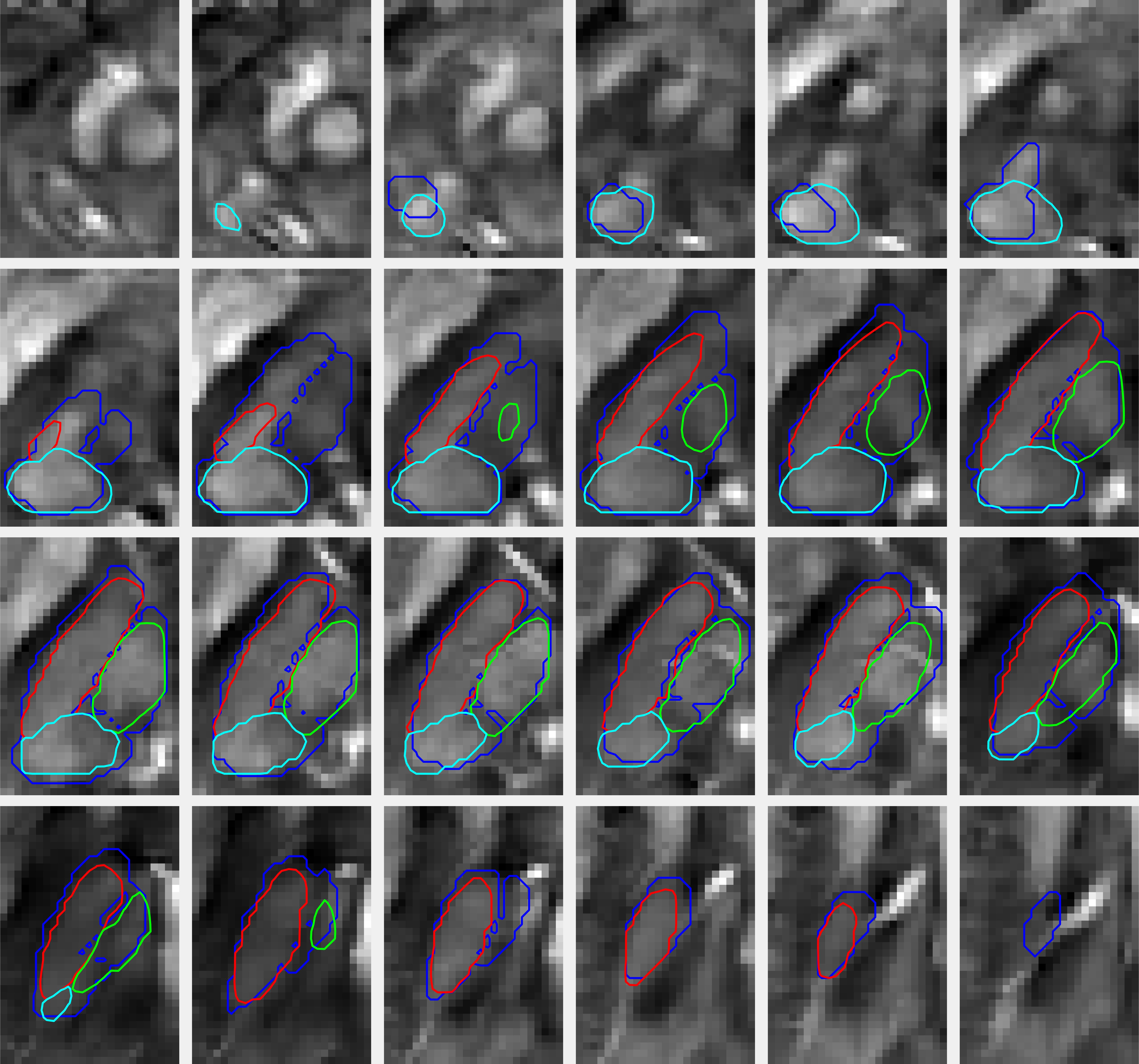}
	\caption[Single-subject segmentation using 3-NN]{Extended view for single-subject thalamus segmentation using 3-NN based pre-processing and $w = 0.4$. The blue overlay represents the ground truth's outer boundary. Red, green and cyan contours are the segmentation results for the three subnuclear groups.}
	\label{fig:s7_04_3nn}
\end{figure}

From a qualitative standpoint, \autoref{fig:single10} illustrates the improvement in single-subject segmentation when weighting the fidelity term by the training-data based prior. Weighted segmentations with both supervised learning approaches (i.e., 3-NN and Parzen) converged to solutions that were overall in close agreement with the manual ground truth. Extended results for the same subject using 3-NN are shown in \autoref{fig:s7_04_3nn}. 
\newline
\newline
\newline

\section{Discussion}
In this study, we present a data-driven method to segment several internal thalamic boundaries using multi-contrast MRI data. We had three imaging contrasts available to drive the segmentation procedure: $T_1$, $T_2^*$-weighting and QSM. We confirmed that using all information maximised segmentation performance, and also found evidence suggesting that QSM was the most informative contrast type for this thalamic problem. Different data types or new implementations for other anatomical regions will require new calibration work.
\newline
\newline
This work was motivated by the observation that study-wise MRI templates obtained using highly iterative non-linear coregistration routines are showing superb anatomical detail over and above what can be inferred from individual datasets. It is therefore unsurprising these are being used to trace regions of interest that are not available from automated segmentation tools \cite{Acosta-Cabronero2016,BETTS201643}. Although this is an effective strategy, it assumes both that there are no co-registration errors in the calculation of the study-wise template space and that the manual reference is an exact definition of the region of interest, which are somewhat inaccurate assumptions. In this work we broke away from this idealisation and propose to correct these errors with two additional steps: one of pattern recognition, followed by convex segmentation promoting (from a Bayesian standpoint) segmentation boundaries that are short, continuous and regular while also consistent with contrast variations in single subjects. Furthermore, to capitalize on the facts (i) that multiple MRI contrasts are typically acquired in the same imaging session, and (ii) that different MRI contrast types could act in concert to help resolve tissue boundaries, the algorithm was implemented in multivariate form. In turn, this new method yielded regional boundaries that were in good agreement with manually traced contours. This was in stark contrast with the output from Morel atlas based segmentation of the same subnuclear groups, confirming that data-driven approaches (such as that which is hereby proposed) signify an improvement (with respect to co-registration based atlas labelling methods) in terms of consistency with manual segmentation.
\newline
\newline
It is also worth noting that posterior probability maps from individual datasets can be noisy. In this study, with a view to make the convex segmentation algorithm more robust in this regime, we introduced an additional data fidelity weight in \eqref{eq:prob_var_single} to enable additional prior knowledge to be transferred from the training reference to single-subject segmentations. Such an approach led to significant improvements for both classifiers, although we noted optimal performance (i.e., lower errors with respect to the gold standard) specifically for 3-NN based modelling and inclusion of 40 \% prior knowledge. Intuitively, $w$-dependent errors reflect the complex interaction between co-registration performance, accuracy on training-template manual delineation and the algorithm's ability to identify biologically meaningful boundaries between tissue types. In other words, the finding that segmentation errors were systematically minimised by $w<1$ confirmed that the proposed algorithm effectively corrects for co-registration and/or manual initialisation errors. We cannot guarantee, however, that the proposed implementation (i.e., 3-NN classification with $\lambda=1$ and CS with $w=0.4$) will be optimal for other regions and/or data types. This warrants further investigation.
\newline
\newline
An additionally important consideration for early adopters of this method is that posterior probability maps can only be obtained from models trained on separate data. In this study, we had sufficient power to split the dataset into training and test subsets. However, future studies wanting to implement this algorithm with limited available data may consider e.g. an algorithmic extension for synthetic data augmentation.
\newline
\newline
In conclusion, this work presented a highly versatile multi-contrast segmentation framework and its successful application to identify thalamic substructures. Future work is warranted to extend this method for segmentation of other structures. In addition to developing appropriate forward pipelines for bootstrapping training data augmentation, further improvements might be obtained using e.g. deep learning, which may eliminate the need for additional spatial constraints.
\newline
\newline
\subsection*{Data statement}
The dataset used in this work and the proposed supervised learning and convex segmentation implementations are available at \url{https://github.com/veronicacorona/multicontrastSegmentation.git}. 

\section*{Acknowledgements}
The Wellcome Centre for Human Neuroimaging is supported by core funding from the Wellcome (203147/Z/16/Z).
VC acknowledges the financial support of the Cambridge Cancer Centre and the Cancer Research UK. CBS acknowledges support from Leverhulme Trust project "Breaking the non-convexity barrier", EPSRC grant "EP/M00483X/1", EPSRC centre "EP/N014588/1", the Cantab Capital Institute for the Mathematics of Information, and from CHiPS and NoMADS (Horizon 2020 RISE project grant). Moreover, CBS is thankful for support by the Alan Turing Institute.

\addcontentsline{toc}{section}{Bibliography}
\bibliographystyle{unsrt}
\bibliography{thesisbiblio}

\begin{thebibliography}{10}

\bibitem{Mai2011}
J{\"u}rgen~K. Mai and George Paxinos.
\newblock {\em {T}he {H}uman {N}ervous {S}ystem}.
\newblock Academic Press, 2011.

\bibitem{Steriade.Llinas}
Mircea Steriade and Rodolfo~R. Llin{\'a}s.
\newblock {T}he functional states of the thalamus and the associated neuronal
  interplay.
\newblock {\em Physiological reviews}, 68(3):649--742, 1988.

\bibitem{Sherman2002}
S.~Murray Sherman and R.~W. Guillery.
\newblock {T}he role of the thalamus in the flow of information to the cortex.
\newblock {\em Philosophical Transactions of the Royal Society B: Biological
  Sciences}, 357(1428):1695--1708, 2002.

\bibitem{Chien2016}
J.~H. Chien, J.~J. Cheng, and F.~A. Lenz.
\newblock {T}he {T}halamus.
\newblock In {\em Conn's Translational Neuroscience}, pages 289--297. Elsevier,
  2016.

\bibitem{Conn2016}
P.~Michael Conn.
\newblock {\em {C}onn's {T}ranslational {N}euroscience}.
\newblock Academic Press, 2016.

\bibitem{Power2015}
Brian~D. Power and Jeffrey C.~L. Looi.
\newblock {T}he thalamus as a putative biomarker in neurodegenerative
  disorders.
\newblock {\em Australian \& New Zealand Journal of Psychiatry},
  49(6):502--518, 2015.

\bibitem{Kassubek.etal}
Jan Kassubek, Freimut~D. Juengling, Daniel Ecker, and G.~Bernhard
  Landwehrmeyer.
\newblock {T}halamic {A}trophy in {H}untington's {D}isease {C}o-varies with
  {C}ognitive {P}erformance: {A} {M}orphometric {M}ri {A}nalysis.
\newblock {\em Cerebral Cortex}, 15(6):846--853, 2005.

\bibitem{Amano2004}
Naoji Amano.
\newblock {N}europsychiatric symptoms and depression in neurodegenerative
  diseases.
\newblock {\em Psychogeriatrics}, 4(1):1--3, 2004.

\bibitem{Ondo2001}
William Ondo, Michael Almaguer, Joseph Jankovic, and Richard~K. Simpson.
\newblock {T}halamic deep brain stimulation: comparison between unilateral and
  bilateral placement.
\newblock {\em Archives of Neurology}, 58(2):218--222, 2001.

\bibitem{Gringel2009}
Tabea Gringel, Walter Schulz-Schaeffer, Erck Elolf, Andreas Fr{\"o}lich, Peter
  Dechent, and Gunther Helms.
\newblock {O}ptimized {H}igh-resolution {M}apping of {M}agnetization {T}ransfer
  (mt) at 3 {T}esla for {D}irect {V}isualization of {S}ubstructures of the
  {H}uman {T}halamus in {C}linically {F}easible {M}easurement {T}ime.
\newblock {\em Journal of Magnetic Resonance Imaging}, 29(6):1285--1292, 2009.

\bibitem{Behrens.etala}
T.~E.~J. Behrens, H.~Johansen-Berg, M.~W. Woolrich, S.~M. Smith, C.~A.~M.
  Wheeler-Kingshott, P.~A. Boulby, G.~J. Barker, E.~L. Sillery, K.~Sheehan,
  O.~Ciccarelli, et~al.
\newblock {N}on-invasive mapping of connections between human thalamus and
  cortex using diffusion imaging.
\newblock {\em Nature neuroscience}, 6(7):750--757, 2003.

\bibitem{Wiegell.etal}
Mette~R. Wiegell, David~S. Tuch, Henrik B.~W. Larsson, and Van~J. Wedeen.
\newblock {A}utomatic {S}egmentation of {T}halamic {N}uclei from {D}iffusion
  {T}ensor {M}agnetic {R}esonance {I}maging.
\newblock {\em NeuroImage}, 19(2):391--401, 2003.

\bibitem{Duan.etal}
Ye~Duan, Xiaoling Li, and Yongjian Xi.
\newblock {T}halamus segmentation from diffusion tensor magnetic resonance
  imaging.
\newblock {\em Int J Biomed Imaging}, 2007:90216, 2007.

\bibitem{Jonasson.etal}
Lisa Jonasson, Patric Hagmann, Claudio Pollo, Xavier Bresson, Cecilia~Richero
  Wilson, Reto Meuli, and Jean-Philippe Thiran.
\newblock {A} {L}evel {S}et {M}ethod for {S}egmentation of the {T}halamus and
  {I}ts {N}uclei in {DT}-{MRI}.
\newblock {\em Signal Processing}, 87(2):309--321, 2007.

\bibitem{Grassi.etal}
A~Grassi, L~Cammoun, C~Pollo, P~Hagmann, R~Meuli, and J~P Thiran.
\newblock {T}halamic nuclei clustering on {H}igh {A}ngular {R}esolution
  {D}iffusion {I}mages.
\newblock {\em Proceedings of the 16th {S}cientific meeting of {I}nternational
  {S}ociety for {M}agnetic {R}esonance in {M}edicine ({ISMRM}) {T}oronto-
  {C}anada {M}ai 2008}, 2008.

\bibitem{Deoni.etal}
Sean C.~L. Deoni, Brian~K. Rutt, Andrew~G. Parrent, and Terry~M. Peters.
\newblock {S}egmentation of thalamic nuclei using a modified k-means clustering
  algorithm and high-resolution quantitative magnetic resonance imaging at 1.5
  {T}.
\newblock {\em Neuroimage}, 34(1):117--126, 2007.

\bibitem{Morel.etal}
Anne Morel, Michel Magnin, Daniel Jeanmonod, et~al.
\newblock {M}ultiarchitectonic and stereotactic atlas of the human thalamus.
\newblock {\em Journal of Comparative Neurology}, 387(4):588--630, 1997.

\bibitem{Magon.etal}
Stefano Magon, M.~Mallar Chakravarty, Michael Amann, Katrin Weier, Yvonne
  Naegelin, Michaela Andelova, Ernst-Wilhelm Radue, Christoph Stippich,
  Jason~P. Lerch, Ludwig Kappos, and Till Sprenger.
\newblock {L}abel-fusion-segmentation and {D}eformation-based {S}hape
  {A}nalysis of {D}eep {G}ray {M}atter in {M}ultiple {S}clerosis: {T}he
  {I}mpact of {T}halamic {S}ubnuclei on {D}isability.
\newblock {\em Hum Brain Mapp}, 35(8):4193--4203, Aug 2014.

\bibitem{Bender.etal}
B.~Bender, C.~M{\"a}nz, A.~Korn, T.~N{\"a}gele, and U.~Klose.
\newblock {O}ptimized 3{D} magnetization-prepared rapid acquisition of gradient
  echo: identification of thalamus substructures at 3{T}.
\newblock {\em American Journal of Neuroradiology}, 32(11):2110--2115, 2011.

\bibitem{Tourdias.etal}
Thomas Tourdias, Manojkumar Saranathan, Ives~R. Levesque, Jason Su, and
  Brian~K. Rutt.
\newblock {V}isualization of {I}ntra-thalamic {N}uclei with {O}ptimized
  {W}hite-matter-nulled {M}prage at 7t.
\newblock {\em Neuroimage}, 84:534--545, 2014.

\bibitem{Deistung.etal}
Andreas Deistung, Andreas Sch{\"a}fer, Ferdinand Schweser, Uta Biedermann,
  Robert Turner, and J{\"u}rgen~R. Reichenbach.
\newblock {T}oward in {V}ivo {H}istology: {A} {C}omparison of {Q}uantitative
  {S}usceptibility {M}apping (qsm) with {M}agnitude-, {P}hase-, and {R} 2
  -imaging at {U}ltra-high {M}agnetic {F}ield {S}trength.
\newblock {\em Neuroimage}, 65:299--314, 2013.

\bibitem{HAMETNER2018117}
Simon Hametner, Verena Endmayr, Andreas Deistung, Pilar Palmrich, Max Prihoda,
  Evelin Haimburger, Christian Menard, Xiang Feng, Thomas Haider, Marianne
  Leisser, Ulrike K{\"o}ck, Alexandra Kaider, Romana Höftberger, Simon
  Robinson, Jürgen~R. Reichenbach, Hans Lassmann, Hannes Traxler, Siegfried
  Trattnig, and G{\"u}nther Grabner.
\newblock {T}he influence of brain iron and myelin on magnetic susceptibility
  and effective transverse relaxation - {A} biochemical and histological
  validation study.
\newblock {\em NeuroImage}, 179:117 -- 133, 2018.

\bibitem{WARD20141045}
Roberta~J Ward, Fabio~A Zucca, Jeff~H Duyn, Robert~R Crichton, and Luigi Zecca.
\newblock {T}he role of iron in brain ageing and neurodegenerative disorders.
\newblock {\em The Lancet Neurology}, 13(10):1045 -- 1060, 2014.

\bibitem{Krauth.etal}
Axel Krauth, Remi Blanc, Alejandra Poveda, Daniel Jeanmonod, Anne Morel, and
  G{\'a}bor Sz{\'e}kely.
\newblock {A} mean three-dimensional atlas of the human thalamus: generation
  from multiple histological data.
\newblock {\em Neuroimage}, 49(3):2053--2062, 2010.

\bibitem{Theodoridis.Koutroumbas}
Sergios Theodoridis and Konstantinos Koutroumbas.
\newblock {\em {P}attern {R}ecognition (fourth {E}dition)}.
\newblock 2008.

\bibitem{Lellmann2011c}
J.~Lellmann and C.~Schn{\"o}rr.
\newblock {C}ontinuous {M}ulticlass {L}abeling {A}pproaches and {A}lgorithms.
\newblock {\em SIAM Journal on Imaging Sciences}, 4(4):1049--1096, 2011.

\bibitem{PockCremersBischofEtAl2009}
Thomas Pock, Daniel Cremers, Horst Bischof, and Antonin Chambolle.
\newblock {A}n algorithm for minimizing the {M}umford-{S}hah functional.
\newblock In {\em Computer Vision, 2009 IEEE 12th International Conference on},
  pages 1133--1140. IEEE, 2009.

\bibitem{ChambolleCremersPock2012}
Antonin Chambolle, Daniel Cremers, and Thomas Pock.
\newblock {A} convex approach to minimal partitions.
\newblock {\em SIAM Journal on Imaging Sciences}, 5(4):1113--1158, 2012.

\bibitem{Acosta-Cabronero2016}
Julio Acosta-Cabronero, Matthew~J. Betts, Arturo Cardenas-Blanco, Shan Yang,
  and Peter~J. Nestor.
\newblock {I}n {V}ivo {MRI} {M}apping of {B}rain {I}ron {D}eposition across the
  {A}dult {L}ifespan.
\newblock {\em Journal of Neuroscience}, 36(2):364--374, 2016.

\bibitem{Folstein.etal}
Marshal~F Folstein, Susan~E Folstein, and Paul~R McHugh.
\newblock "{M}ini-mental state": a practical method for grading the cognitive
  state of patients for the clinician.
\newblock {\em Journal of psychiatric research}, 12(3):189--198, 1975.

\bibitem{MuglerIII1990}
John~P Mugler~III and James~R Brookeman.
\newblock {T}hree-dimensional magnetization-prepared rapid gradient-echo
  imaging (3{D} {MP} {RAGE}).
\newblock {\em Magnetic Resonance in Medicine}, 15(1):152--157, 1990.

\bibitem{Wang.Liu}
Yi~Wang and Tian Liu.
\newblock {Q}uantitative {S}usceptibility {M}apping (qsm): {D}ecoding {M}ri
  {D}ata for a {T}issue {M}agnetic {B}iomarker.
\newblock {\em Magnetic Resonance in Medicine}, 73(1):82--101, 2015.

\bibitem{Tustison.etal}
Nicholas~J Tustison, Brian~B Avants, Philip~A Cook, Yuanjie Zheng, Alexander
  Egan, Paul~A Yushkevich, and James~C Gee.
\newblock {N}4{ITK}: improved {N}3 bias correction.
\newblock {\em Medical Imaging, IEEE Transactions on}, 29(6):1310--1320, 2010.

\bibitem{ants}
Brian~B. Avants, Nick Tustison, and Gang Song.
\newblock {A}dvanced {N}ormalization {T}ools ({ANTS}).
\newblock {\em Insight J.}, 2009.

\bibitem{Avants.etal}
Brian~B. Avants, Charles~L. Epstein, Murray Grossman, and James~C. Gee.
\newblock {S}ymmetric diffeomorphic image registration with cross-correlation:
  evaluating automated labeling of elderly and neurodegenerative brain.
\newblock {\em Medical image analysis}, 12(1):26--41, 2008.

\bibitem{BETTS201643}
Matthew~J. Betts, Julio Acosta-Cabronero, Arturo Cardenas-Blanco, Peter~J.
  Nestor, and Emrah Düzel.
\newblock {H}igh-resolution characterisation of the aging brain using
  simultaneous quantitative susceptibility mapping ({QSM}) and {R}2*
  measurements at 7{T}.
\newblock {\em NeuroImage}, 138:43 -- 63, 2016.

\end{thebibliography}

\end{document}